\documentclass[10pt,letterpaper]{article}
\usepackage[top=0.85in,left=2.75in,footskip=0.75in]{geometry}

\usepackage{amsmath,amssymb}

\usepackage{changepage}

\usepackage{textcomp,marvosym}

\usepackage{cite}

\usepackage{nameref,hyperref}

\usepackage[right]{lineno}

\usepackage[nopatch=eqnum]{microtype}
\DisableLigatures[f]{encoding = *, family = * }

\usepackage[table]{xcolor}

\usepackage{array}
\usepackage{amsmath}
\newcolumntype{+}{!{\vrule width 2pt}}

\newlength\savedwidth

\newcommand\thickhline{\noalign{\global\savedwidth\arrayrulewidth\global\arrayrulewidth 2pt}%
\hline
\noalign{\global\arrayrulewidth\savedwidth}}

\raggedright
\usepackage[aboveskip=1pt,labelfont=bf,labelsep=period,justification=raggedright,singlelinecheck=off]{caption}

\makeatletter
\renewcommand{\@biblabel}[1]{\quad#1.}
\makeatother

\usepackage{lastpage,fancyhdr,graphicx}
\usepackage{epstopdf}
\fancyheadoffset[L]{2.25in}
\fancyfootoffset[L]{2.25in}
\begin{document}
\vspace*{0.2in}

\begin{flushleft}
{\Large
\textbf\newline{FaceTouch: Detecting hand-to-face touch with supervised contrastive learning to assist in tracing infectious diseases} 
}
\newline
\\
Mohamed R. Ibrahim\textsuperscript{*,1,2,3},
Terry Lyons\textsuperscript{3,4}

\bigskip
\textbf{1} Centre for Data Analysis and Policy, University of Leeds, Leeds, UK
\\
\textbf{2} Leeds Institute for Data Analytics (LIDA)
\\
\textbf{3} The Alan Turing Institute, London, UK
\\
\textbf{4} Mathematical Institute, University of Oxford 
\\
\bigskip

%
%


\textcurrency Current Address: Department of Environment, School of Geography, Centre of Data Analysis and Policy, University of Leeds, Leeds, UK 



* geomi@leeds.ac.uk

\end{flushleft}
\section*{Abstract}
Through our respiratory system, many viruses and diseases frequently spread and pass from one person to another. Covid-19 served as an example of how crucial it is to track down and cut back on contacts to stop its spread. There is a clear gap in finding automatic methods that can detect hand-to-face contact in complex urban scenes or indoors. In this paper, we introduce a computer vision framework, called FaceTouch, based on deep learning. It comprises deep sub-models to detect humans and analyse their actions. FaceTouch seeks to detect hand-to-face touches in the wild, such as through video chats, bus footage, or CCTV feeds. Despite partial occlusion of faces, the introduced system learns to detect face touches from the RGB representation of a given scene by utilising the representation of the body gestures such as arm movement. This has been demonstrated to be useful in complex urban scenarios beyond simply identifying hand movement and its closeness to faces. Relying on Supervised Contrastive Learning, the introduced model is trained on our collected dataset, given the absence of other benchmark datasets. The framework shows a strong validation in unseen datasets which opens the door for potential deployment.



\section*{Introduction}
Humans have an innate habit of touching their faces \cite{1,2}. Touching sensitive/mucosal face zones (eyes, nose, and mouth) frequently increases health risks by introducing microorganisms into the body and spreading disease \cite{3,4}. In addition, reliable monitoring of facial touch is required for behavioural intervention. Building an automated system that can understand human behaviour in complex environments is crucial for many applications. In times of pandemics, detecting and tracing where our hand touches could lead to a better understanding of how infectious viruses spread. 

In recent years, computer vision and deep learning have made significant progress in comprehending numerous aspects related to human actions and their perception and interaction with the built environment \cite{5,6,7,8,9}. While there is still a clear gap in finding real-world image datasets for recognising hand-to-face touch \cite{10}, there are some works that focus on detecting face touch by relying on smart devices worn by participants, which makes it challenging and unsustainable system to use to understand human movements as it requires multiple data sources from individuals. On the other hand, other systems have been developed relying on independently detecting hands and faces and classifying the hand-to-face touch based on the proximity of one to the other. As a result, when detecting hand or object movement close to one's face, these systems are more likely to have flaws and a higher likelihood of false positives (for example, drinking from a water bottle, or picking up a phone).

In this research, we contribute with the following:  1) We introduce a new framework called FaceTouch that aims to detect hand-to-face touches in the wild, including video calls, bus footage, or CCTV feeds. The introduced framework learns to detect face touches from the RGB representation of a given scene, despite partial occlusion of faces. It made use of body gestures such as the movement of the arm which is useful in complex urban scenes, beyond detecting hand movement and its proximity to faces alone. 2) We extend the widely utilised self-supervised batch contrastive learning to fully-supervised learning, allowing us to effectively exploit image labels. 3) We also introduce a new dataset for face-to-hand touch including various poses of humans at both indoor and outdoor,  last, 4) we also provide an extensive analysis of different deep learning models that can contribute to solving other similar issues.

After the introduction, section 2 describes the related work and methods previously used for the stated issues. Section 3 describes the introduced framework, training procedures and evaluation metrics. Section 4 summarises the results of the method, and section 5 discusses the results with the current literature, highlighting future work and limitations. Last, we conclude our work.

\section*{Related work}

Some studies in the literature can be directly linked to the stated issues, which can be summarised in two domains.  

\textbf{Detection via sensor devices: }By relying on smartwatches as data sensors, \cite{11} developed a method to detect spontaneous facial self-touches by extracting and classifying accelerometer data from various participants via various machine learning methods, including Random Forest and Support Vector Machines. In a similar approach to analysing hand movement,  \cite{3} used accelerometer data to detect face touches using Random Forest. Similarly, A wearable system has been introduced to avoid unconscious face touches relying on accelerometer data and a deep learning approach to classify hand movement \cite{4}.  On the other hand, \cite{2} used an ear-worn device and developed a method to detect hand touches and identify the types of hand touches to mucosal and non-mucosal areas by relying on thermal infrared and physiological signals determining changes caused by the skin when a face is touched. 

\textbf{Separate hand and face detections: }Despite accuracy, detecting a facial touch can be achieved by detecting individually one’s hand and face and arithmetically finding the threshold distance to determine a touch.  First, regarding face detection, several works have been achieved to detect faces as a lightweight method that can be used for edge devices in real-time \cite{12,13,14}. Furthermore, Deng et al. \cite{15} developed a method for detecting faces and key facial landmarks in the wild by relying on feature pyramids and deep architecture. This introduced method shows a strong approach for detecting and localising a large number of faces in a given image. Similarly, Hu and Ramanan \cite{11} developed a method to detect tiny faces in the wild relying on CNN architecture and re-scaling the input image to different sizes allowing the introduced method to detect faces simultaneously at a different resolution to output the final detection of the merged outcomes of each resolution. Furthermore, Yang and Song \cite{17} introduced a new loss function for deep learning that could enhance facial recognition in different illumination settings.  Second, regarding hand detection, Adiguna and Soelistio \cite{18} used the CNN model to create a posture-free hand detector from RGB images. Liu et al. \cite{13} extended the detection of hand from pixels by introducing deep blocks that can allow better interpretation of the results and a robust rotation map that provides a rotated bounding box of hands in the wild, similar to the work developed by \cite{20}. Moreover, Xu et al. \cite{21} developed a method to detect hands by reconstructing their representation using Generative Adversarial Networks (GAN).  In contrast, Kourbane and Genc \cite{22} developed a skeleton-aware regressor model to estimate 2D hand pose by relying on the key points of hands. Furthermore, different research has been developed to provide a detailed detection of hand gestures to perform complex tasks \cite{23,24,25}. 

In summary, progress has been made by using data from sensing devices to categorise hand motion and, consequently, define a hand-to-face touch. Additionally, an arithmetic method can be used to calculate the distance between a hand and a face to identify a touch by localising both hands and faces from images. On the other side, a system to localise face occlusion by hands was developed \cite{10},  acquiring the ability to recognise hand-over-face occlusions by synthesising facial occlusions from a dataset of non-occluded faces. However, difficulties still exist in overcoming the shortcomings of the aforementioned methodologies and learning to comprehend a face touch through the representation of faces or human poses in the wild (not necessarily where all faces or hands can be seen).

\section*{Materials and methods}
The project was approved ethically by Urban Observatory, New Castle University, where the funds are allocated. Individual consent is not required because the data utilised is not disclosed and includes no personal information. We only reveal the findings based on publicly available internet data, with blurry faces. 

In this section, we describe the approach to our method, architecture, materials,  evaluation metrics, and implementation details, including model hyperparameters.

\subsection*{Approach}

To detect a face touch based on an input of a given RGB image at different scales and high variance whereas a given face can be seen clearly (i.e. at a video call), or at very low resolution with many occlusions (i.e. CCTV cameras in street), we relied on supervised contrastive learning (SCL) approach \cite{26}. SCL is closely linked to Triplet loss \cite{27,28}, proven to be superior in performance in comparison to the traditional approach of supervised learning.

 In SCL, a model learns a given task through two networks; 1) An encoder network [Enc(.)] and  2) a projection network $[Proj(.)]$. First, the encoder network maps x to a vector representation  r, given that $r=Enc(x)\in \mathbb{R}^{D_E}$, whereas a given sample $x$ is augmented by finding a random different view of the sample and passed to the model paired with the original one. Any known image recognition architecture (i.e. ResNets, MobileNet) could be used to represent this encoder network. Second, the projection network maps $r$ to a vector $z$, given that $z=Proj(r) \in \mathbb{R}^{D_P}$. This enables using the inner product to measure distances in the latent vector space. The architecture of this encoder could be a single fully-connected layer or a multi-perception layer.  It is worth mentioning that The projection network is used only during training whereas it is discarded during inference, making the inference time dependent only on the architecture of the encoder network.

\begin{figure}[ht]
\includegraphics[width=1\linewidth]{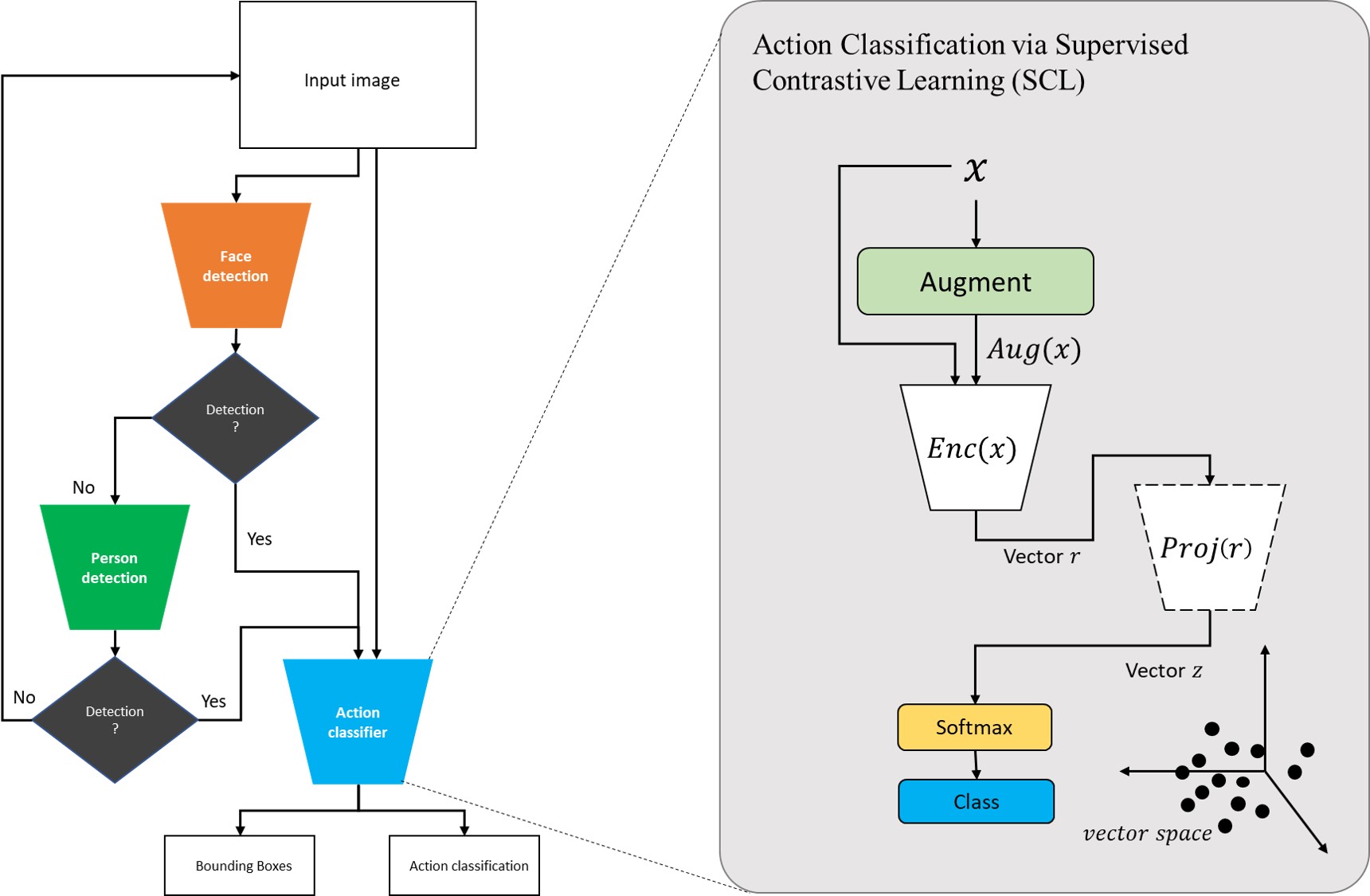}
\caption{{\bf The architecture of FaceTouch.}
}
\label{fig1}
\end{figure}

\subsection*{Proposed Framework}
To utilise the learned model to be used in practice with for example a stream of video data, we have created a framework that allows the detection and localisation of humans and faces at different scales. Fig \ref{fig1} shows the overall framework comprising  four main components as follows: 

\textbf{Backbone:}To determine humans, we rely on two backbone models for the introduced method including, object detection and face detection to maximize the classification of a face touch in complex scenes. Initially, a face detector is used to detect faces, while the human detector is inactive, in the case of face detection, the subset of images that includes faces are passed to the encoder to detect hand action. In the case of undetected faces, a human detector is activated to detect humans. In case of a positive detection, the sub-set of a given scene that includes humans are passed to the following encoder to detect hand actions. The architecture of the object detection is based on YoloV5, which is proven to achieve high results in the VOC dataset \cite{29}, in comparison to other state-of-the-art methods besides its real-time performance. This trade-off between performance and speed allows using the proposed object detector as a backbone. As for face detection, we utilised Haar Cascades algorithms \cite{30}, which have been also proven to be a fast and simple approach for detecting faces with minimal computational needs.

\textbf{Action encoder:} After the introduced backbone, the extracted faces/bodies RGB images are passed through an encoder to classify them and identify whether there is a face touch while localising multi-humans in a given image. As explained previously, we utilise here SCL approach to learn a face touch. Nevertheless, to optimise speed and efficiency for the overall framework,  we have trained several state-of-the-art encoder architectures, including ResNet \cite{31}, MobileNet \cite{32,33}, ImageSig \cite{34}, and Vision Transformer (ViT) \cite{35} with both approaches: traditional supervised learning and SCL.  Accordingly, an optimal architecture and learning approach can be selected. We report on how each model is trained in the implementation section and we provide a quantitative analysis of the pre-trained models on the introduced validation dataset, in the results section. 

\textbf{Face Blur:} To comply with data privacy, ethical considerations, and minimising the re-identifications of human subjects, in the case of face detection, we provide a component that adds Gaussian noise to the local distribution of a given image to ensure data anonymity when using the tool in practice.  

\textbf{Explainable AI:} We also added a component to visualise attention when inferring face. We have used Grad-CAM \cite{36} to visualise the learned weights and localise the attention when classifying a hand touch in the trained model. This implementation occurs only after training at the inference, where this strategy has demonstrated effective results in the goal of our model for displaying context-aware and localised attention with a small number of parameters.

\subsection*{Framework losses and evaluation metrics}
For the introduced backbone of object detection to detect and localise humans, we define the  object loss based on the weighted sum of the localization loss $( L_loc)$ and  confidence loss $(L_conf)$, as follows: 

\begin{eqnarray}
\label{eq:losses}
	\mathrm{L(x,l,g)} =  \frac{1}{N} (L_{conf} (x,c)+\alpha L_{loc} (x,l,g)),
\end{eqnarray}
Given that $N$ represents the matched default bounding boxes, the loss is set to $0$ if $N=0$ and  $\alpha$ is set to $1$ by cross-validating the model. The confidence loss represents a cross-entropy loss based on a softmax loss for the different classes $(c)$.  The parameters of the predicted box $(l)$ are defined based on the default bounding box's centre $(cx,xy)$, as well as its width $(w)$ and height $(h)$ and the localization loss is described as a smooth loss between those parameters and the ground truth bounding box $(g)$. It is defined as:

\begin{eqnarray}
\label{eq:localisation}
	\mathrm{L_{loc}(x,l,g)} = \sum_{i \in Pos}^{N} \sum_{m \in {cx,xy,w,h}} x_{ij}^k  smooth L_{1} (l_i^m-\hat{g}_i^m),
\end{eqnarray}
To train the action encoder for hand-to-face touches based on supervised contrastive learning, whereas there are more negative cases than positive, we utilise a supervised contrastive loss defined as:

\begin{align}
\label{eq:contrative1}
\mathrm{L_{Contrastive}} = \sum_{i\in I} - \log \left( \frac{1}{|P(i)|} \sum_{p\in P(i)} \frac{\exp(z_i \cdot \frac{Z_p}{\tau})}{\sum_{a\in A(i)} \exp(z_i \cdot \frac{Z_a}{\tau})} \right)
\end{align}

where $Z_l$ represents the projection of $Enc(\widetilde{x_l})$, $\tau$ is a parameter for scalar temperature, and represents the anchor,  given that$ P\left(i\right)\equiv\left\{p\in A\left(i\right):\widetilde{y_p}=\widetilde{y_i}\right\}$ represents the set of indices of the positive cases in a given batch of $(i)$ and is $|P(i)|$ the cardinality of $P(i)$.

As a baseline, we also used a  traditional cross entropy function paired with a focal loss to account for the presented class imbalance in the data sets with skew representation for each class. Providing that $y\in{0,1}$ and $\hat{P}\in{0,1}$, the objective loss $L$ is defined as: 

\begin{eqnarray}
\label{eq:loss3}
\mathrm{L(y,\hat{P})}=-\alpha y(1-\hat{P})^{\gamma} \log \hat{P} -(1-y)^{\hat{P}\gamma}log(1-\hat{P})
\end{eqnarray}
given that $\gamma$ is a focusing parameter that specifies how much higher-confidence correct predictions contribute to the overall loss (the higher  $\gamma$, the faster easy-to-classify examples are down-weighted), and $\alpha$  is a hyperparameter that governs the trade-off between precision and recall by weighting errors for the positive class up or down $(\alpha =1)$ is the default, which is the same as no weighting). 

To evaluate  the performance of the trained models based on different approaches with different encoder’s architecture, we calculated accuracy as $(TP+TN)/(TP+TN+FP+FN)$, precision as$ TP/(TP+FP)$, recall as $TP/(TP+FN)$, and F1-score $2\times\frac{Precision\times R e c a l l}{Precision+Recall}$ , given that $TN$ are the predicted true-negative values, $TP$ are the predicted true-positive values, $FN$ are the predicted false-negative values and  $FP$ are the predicted false-positive values. We also computed the Receiver operating characteristic curve (ROC) curve to outline the performance of the classification for both: the backbone and the action encoders.

\subsection*{Materials}

\begin{figure}[ht]
\includegraphics[width=1\linewidth]{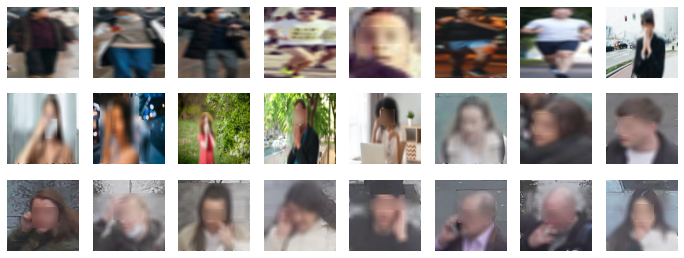}
\caption{{\bf A sample of the collected data that belongs  (Faces are blurred for privacy).}
}
\label{fig2}
\end{figure}

There are no open-access deep learning datasets that label and classify persons in complex settings based on whether one touches their own face. Given this limitation and the lack of a benchmark data set, compiling our database becomes a reliable resource for carrying out this research. We gathered over 20,000 images from the Google Images search engine that relate to indoor and outdoor locations, including diverse positions of humans in various settings and conditions. Furthermore, these images are collected without regard to image size, the presence of urban structures, components, or field of view. Following a visual inspection, we focused on 10,413 images of individuals touching or not touching their faces. These images are divided according to training and test with 0.8 to 0.2 proportions, respectively.  Fig \ref{fig2} shows a sample of the collected data for people self-touching their faces and without touching their faces. The images belong to a different context and various resolutions to ensure learning the complexity of a face-touch in a real-world setting. 

Three criteria are used to define the model's ground truth. First, there is the obvious case of people touching their faces. Second, when data is acquired, the metadata linked with the images from search engines (such as Google Search engine) is used. In all, the collected images are visually inspected to ensure label relevancy and a broad representation of image orientation and illumination conditions, and only then are the images labelled to either face touch or not. 

We trained the model to interpret both facial images and whole body images to enable a higher degree of freedom in analysing the condition of a wide range of human movements and gestures within given scene images. As a result, regardless of the angle or elevation of the input image, the model will be able to recognise the status of hand touch. While this change complicates the training process, it enables the proposed model to be extensively used and subsequently enhanced to fulfil a variety of sensing needs for indoor and outdoor settings. Accordingly, We also used the VOC benchmark dataset   \cite{29} to train an object detector to localise humans. Table \ref{table1} summarises the two datasets used in this research, in terms of size and classes. 

\begin{table}[ht]
\centering
\caption{
{\bf The datasets used for different tasks}}
\begin{tabular}{|l+l|l|l|l|}
\hline
\bf Task &	Data set name & Size &  classes \\ \thickhline

Action recognition	& FaceTouch (New)	& 10,413	& Negative: 8,101, positive: 2,312 \\ \hline

Object detection	& VOC  (Benchmark)	& 11,540	&  A multi-class dataset t\\ \hline

\end{tabular}

\label{table1}
\end{table}

\subsection*{Implementation details}

\textbf{Object detection: }For the backbone of the introduced framework, we trained an object detector on the VOC dataset based on the architecture of YOLOV5l. We followed closely the training procedures introduced for YOLOV5.  First, the anchor size has been self-calibrated on the training set of the dataset, resulting in a size of 4. The model is optimised based on a learning rate of 0.01, and a momentum of 0.937 with a weight decay of 0.0005. We used several data augmentation techniques, including translation, scale, shear, flipping, and mosaic techniques. We trained the model with a pixel size of 640 and with a batch size of 4 for 100 epochs. 

\textbf{Action recognition: }For classifying actions, we trained several classifiers following two different training procedures as introduced early (with contrastive loss and with traditional cross entropy/focal losses) with different architectures to account for the trade-off between speed and accuracy. First, for traditional supervised learning, we trained architecture such as ImageSig by following closely the implementation of ImageSig as presented.  Following signature computation, we employed a basic convolution block composed of two CNN 1D layers with feature maps of 32 and 64 and a kernel size of 3, all of which are activated by a ReLU function. A Max-pooling layer of kernel size 3 follows each layer. The model is flattened after the last pooling layer and feedforward to a single 50-neuron FC layer that is activated by a ReLU function before the last softmax layer. The Adam optimization method is used to train the model, using a batch size of 3000 for 300 iterations. We also trained Convolutional models by utilising transfer learning to train the presented convolution-based models, based on ImageNet weights. For training ResNet models \cite{31} and MobileNet models, After freezing the model’s weight, We trained an FC and output layer using the same hyperparameters as the aforementioned architectures after truncating the FCs layer in each given model. We noticed that models converged in the given dataset when trained with a batch size of 32 and for 50 epochs. We trained the Vit model using an image patch size of 6 and an input size of 64 X 64. The architecture consists of four transformer layers with 64 projection dimensions. Each layer has four attention heads and transformer units of 128, and 64 respectively.  The model is trained using the AdamW optimizer with 256 batches and 20 epochs. 

Second, training the SCL approach is slightly different than a traditional supervised classifier. It includes training a given architecture in two stages. First, we started with unsupervised training of the encoder with the Proj(.) network based on the introduced contrastive loss. Afterward, the weights of this network are frozen and re-trained with supervision on the data labels using the Enc(.) network. We trained each introduced architecture followed by a single FC layer of projection units of 128, in the case of the Proj(.) network activated by a ReLU function, dropout rate of 0.5, temperature value of 0.05. Whereas in the case of the  Enc(.) network, we added a hidden layer of 512 neurons and activated also with a ReLU function and followed by an output layer of the binary classes activated by a Softmax layer. All models followed the same optimisation procedures based on Adam optimiser with a learning rate of 0.001 and momentum of 0.9 and trained with 256 batch size for 50 epochs.

\section*{Results and ablation study}

\begin{table}[!ht]
\begin{adjustwidth}{-2.25in}{0in} 
\centering
\caption{
{\bf Evaluation metrics of the trained models of the FaceTouch framework}}
\begin{tabular}{|l+l|l|l|l|l|l|l|l|}
\hline
\bf Task & Models & Size & Acc. (\%)	& AP	& Recall	& F1-score	& Params (m)	 & Size (MB)
 \\ \thickhline

Backbone & YoloV5l	& (640,640)	& -	& 0.490$^1$	& 0.732$^1$	& 0.570$^1$	& 46.5 &	14.0
 \\ \hline

Action Recognition: SL & ResNet50v2	& (64,64)	& 93.839	& 0.801	& 0.891	& 0.879	& 23.6	& 102.0\\

 & MobileNet	& (64,64)	& 95.929	& 0.864	& 0.934	& 0.920	& 3.2	& 18.5\\
& MobileNetv2	& (64,64)	& 94.059	& 0.808	& 0.878	& 0.882	& 2.2	& 16.4\\

&VGG16 &	(64,64)	& 90.319	& 0.708 &	0.886	& 0.822	& 14.7	& 59.2\\

& SigModel$^2$&  (64,64)	& 84.268	& 0.563	& 0.751	& 0.706	& 0.037	& 0.48\\ 

& ViT &  (64,64)	& 91.24 & 0.973  & 0.889	& 	 0.940 & 	7.9 &95.5 \\ 

\hline

Action Recognition: SCL & ResNet50v2	& (64,64)	& 96.920	& 0.897	& 0.939	& 0.939	& 23.6	& 102.0 \\

& MobileNet	& (64,64)	& 98.020	& 0.931	& 0.974	& 0.961	& 3.2	& 18.5 \\

& MobileNetv2	& (64,64)	& 96.040	& 0.867 & 0.943	& 0.923	& 2.2	& 16.4 \\

& VGG16	& (64,64)	& 96.590	& 0.883	& 0.965	& 0.934	& 14.7	& 59.2\\ \hline

\end{tabular}
\begin{flushleft} 
$^1$Represents Mean values for AP, Recall, and F1-score. 

$^2$The model is trained with a truncated signature depth (N) of 4. 

\end{flushleft}

\label{table2}
\end{adjustwidth}
\end{table}

After training the different models in the framework of FaceTouch. Table \ref{table2} shows the evaluation metrics of the different models in two different approaches, including traditional supervised learning and supervised contrastive learning approaches. It shows a substantial performance improvement when it comes to SCL for each given architecture network. For instance, training a network based on VGG16 architecture has improved by 5.2\% in comparison to the traditional supervision methods. The table also shows not only the accuracies of each model but also other metrics, including AP, Recall and F1-score, in addition to the size of each one to evaluate the trade-off between accuracies, model stability in classification, and their size within the overall framework.

\begin{figure}[ht]
\includegraphics[width=1\linewidth]{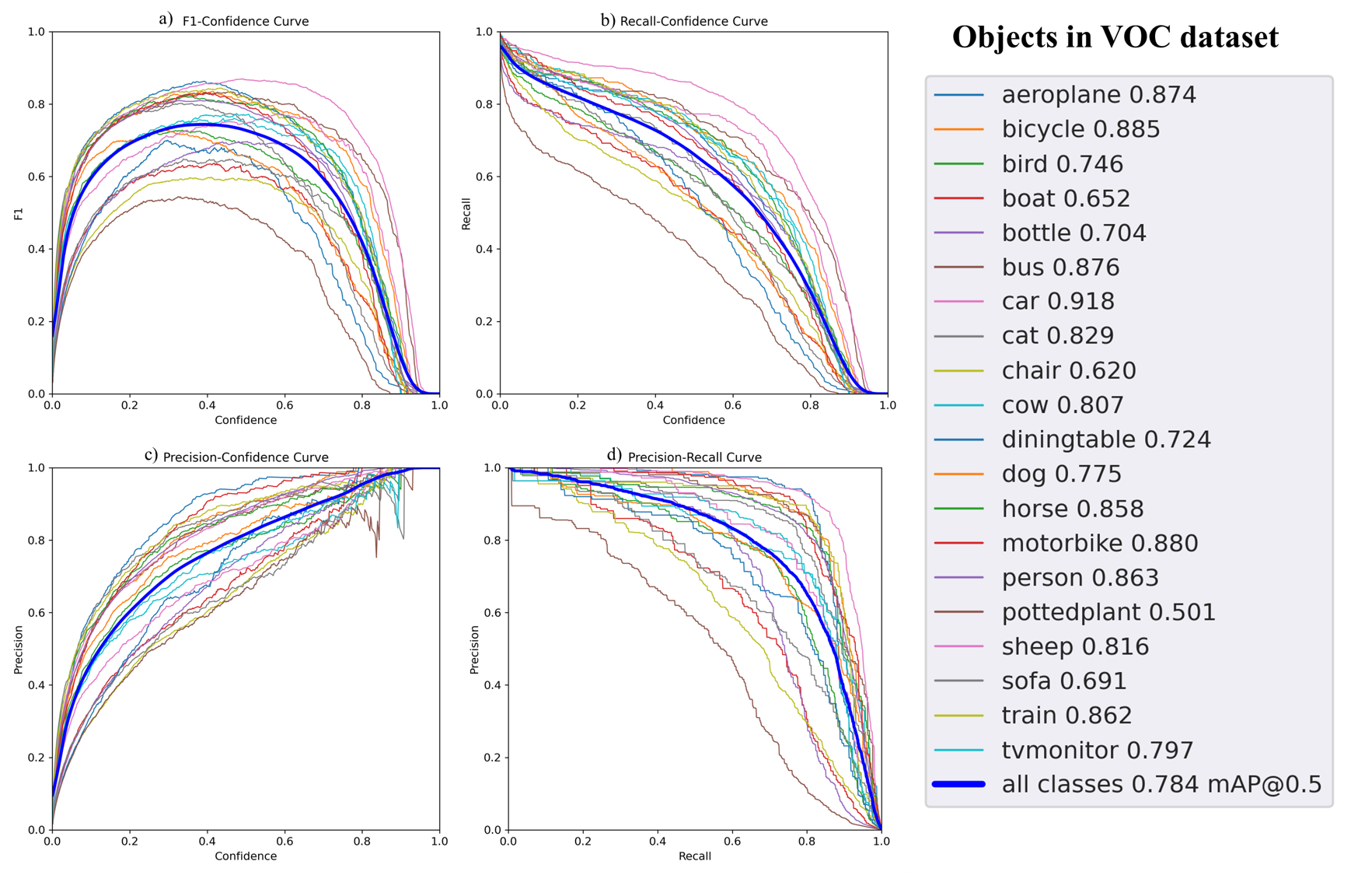}
\caption{{\bf Evaluation of the object detection model.}  (a) describes the relationship between F1 and confidence for the different classes of the model. (b) describes the relationship between Recall and confidence for the different classes of the model. (c) describes the relationship between Precision and confidence for the different classes of the model. (d) describes the relationship between Precision and Recall for the different classes of the model, highlighting the average curve (in blue colour). 
}
\label{fig3}
\end{figure}

\begin{figure}[ht]
\includegraphics[width=1\linewidth]{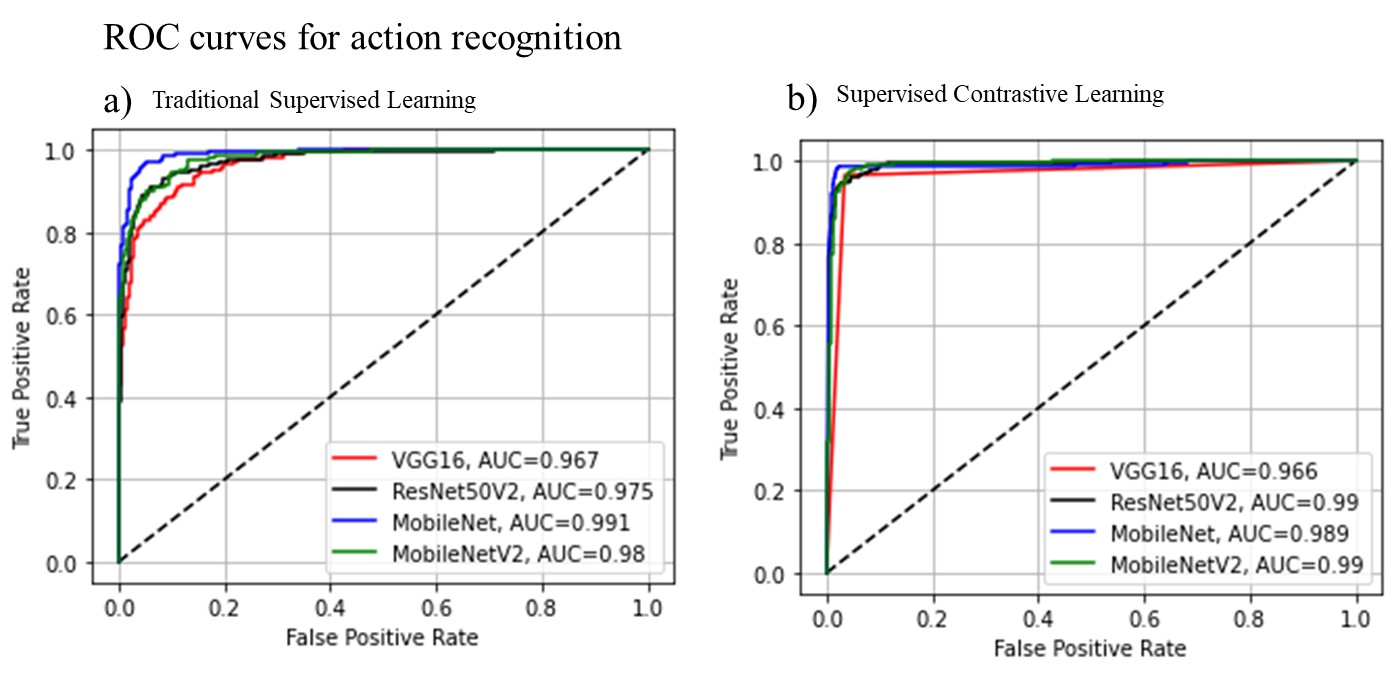}
\caption{{\bf ROC curves for trained action recognition models.}  (a) represents the trained models with supervised learning. (b) represents the trained models with Supervised Contrastive Learning. 
}
\label{fig4}
\end{figure}

Fig \ref{fig3} shows the evaluation metrics for the object detection model. It highlights the relationships of F1, confidence curve, precision and recalls for all classes in the VOC dataset, showing high performance for human detection across the different metrics. Fig \ref{fig4} shows the ROC curves for all trained action recognition, including SCL and traditional supervised learning. Without a doubt, training the all presented architectures with SCL outperforms the traditional approach.

\begin{figure}[ht]
\includegraphics[width=1\linewidth]{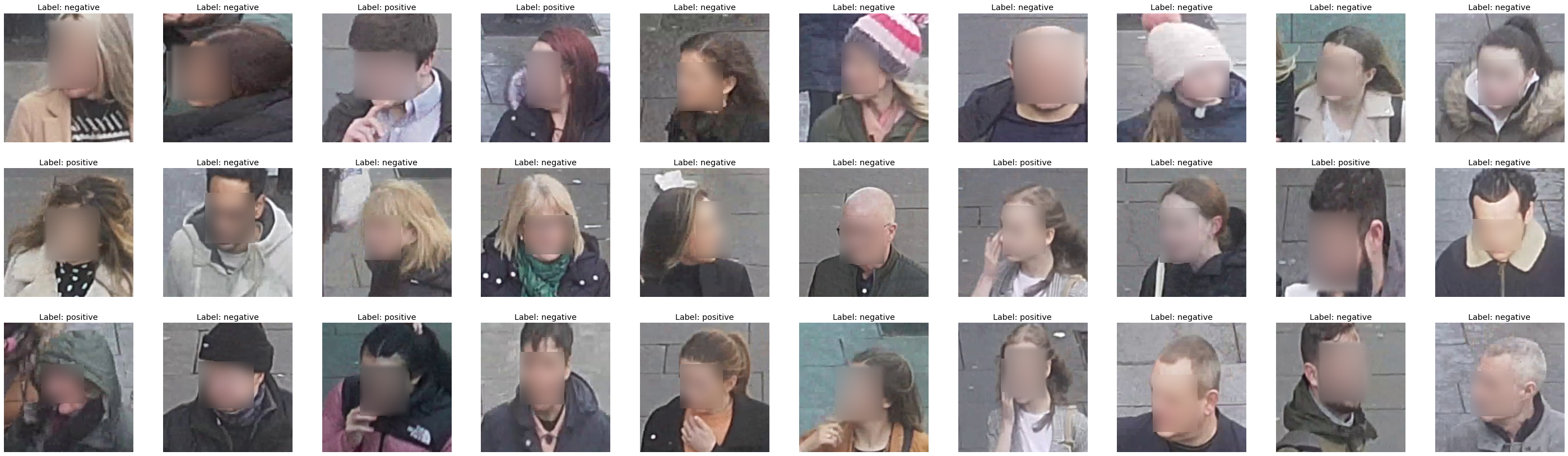}
\caption{{\bf Examples of the predicted positive and negative cases for face touches.} 
}
\label{fig5}
\end{figure}

\begin{figure}[ht]
\includegraphics[width=1\linewidth]{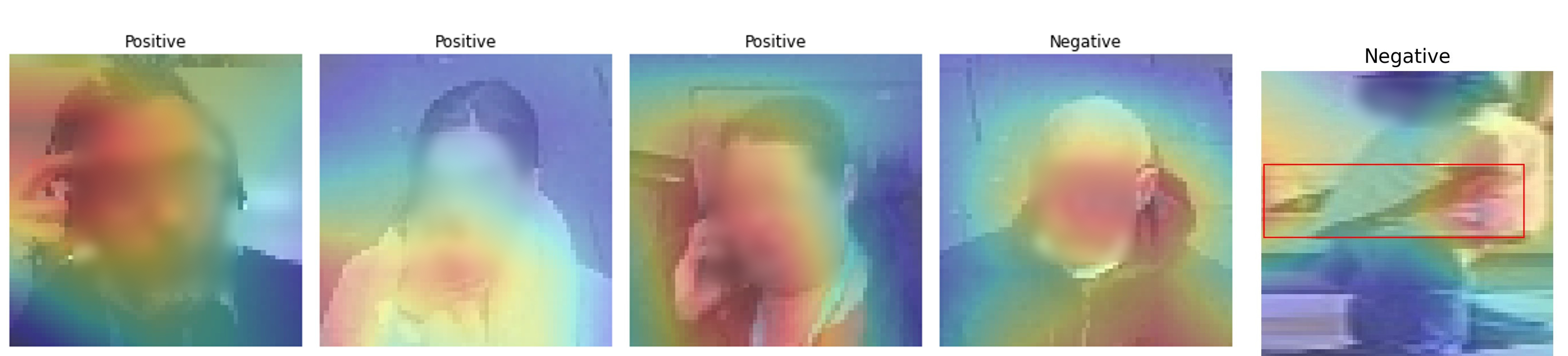}
\caption{{\bf Examples of overlaying the learned attention of the model with the images, highlighting a high accuracy of localising the attention on the faces and hands.} 
}
\label{fig6}
\end{figure}

\begin{figure}[ht]
\includegraphics[width=0.7\linewidth]{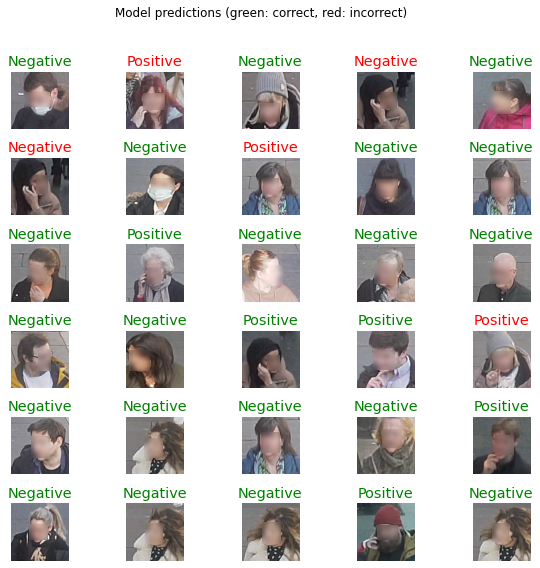}
\caption{{\bf Examples of s incorrectly identified cases (highlighted in red) in comparison to correctly labelled ones (highlighted in green).} 
}
\label{fig6-2}
\end{figure}

Fig \ref{fig5} shows a sample of images in the test set, highlighting a wide range of scene contexts where the model succeeded to classify a hand touch. To verify what the model has learned,  Fig \ref{fig6} shows the weights of the trained model before classification. It explains how the model concludes the classification of a given class. It shows the strength of the model in self-localising and focusing its weights on hand position, when hands are available in a given scene, and face.  Fig \ref{fig6-2} shows incorrectly identified cases in contrast to successfully labelled ones, particularly when the hand is close to the side of the face, suggesting that more attention is needed to further improve inference in this case. 

\section*{Discussion}

\begin{figure}[!h]
\begin{center}
\includegraphics[width=0.6\linewidth]{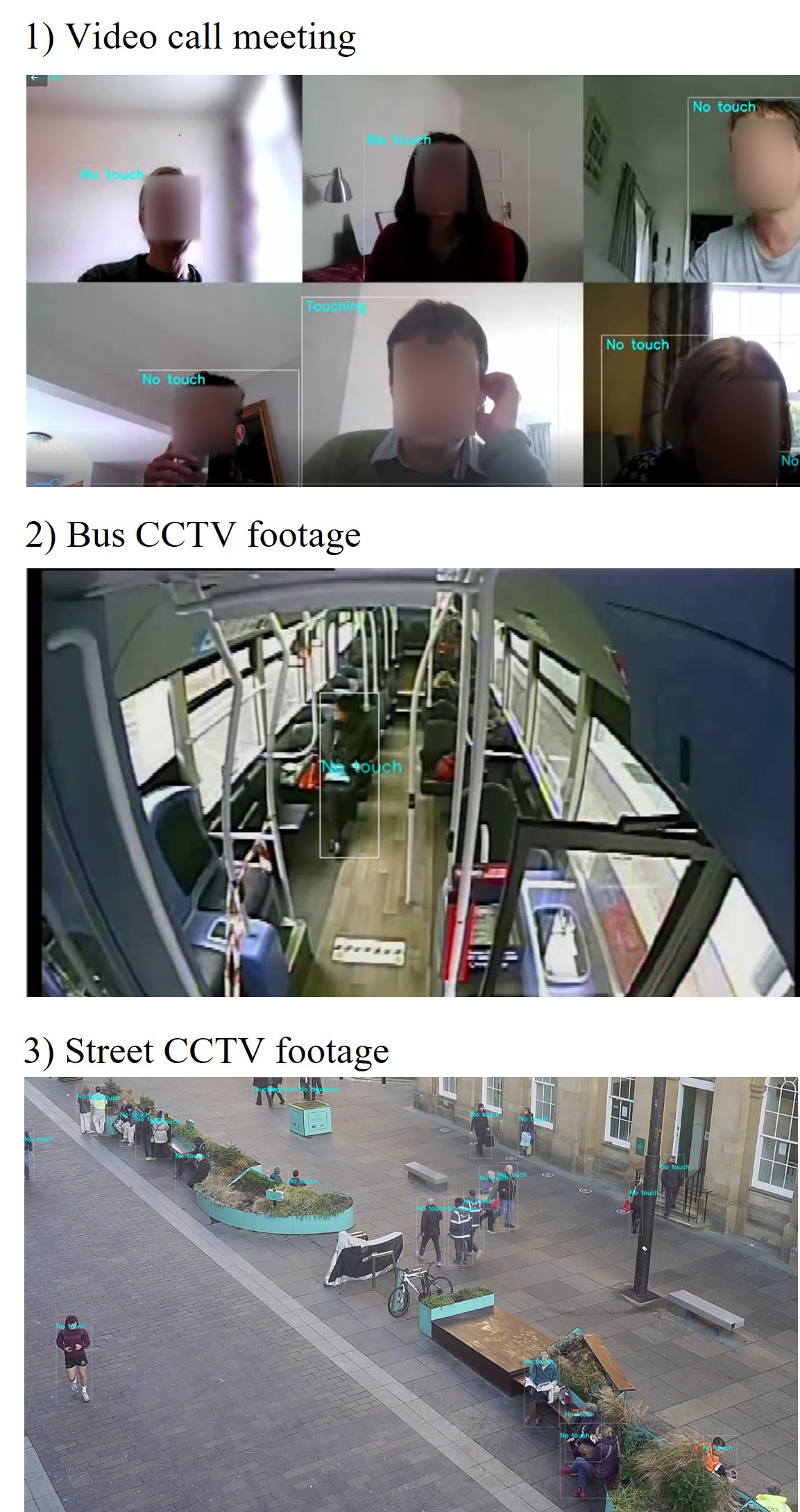}
\caption{{\bf Deploying the FaceTouch tool in video streams of several complex settings such as video calls, bus CCTV footage, and street CCTV.} 
}
\label{fig7}
\end{center}

\end{figure}

\begin{figure}[!h]
\includegraphics[width=1\linewidth]{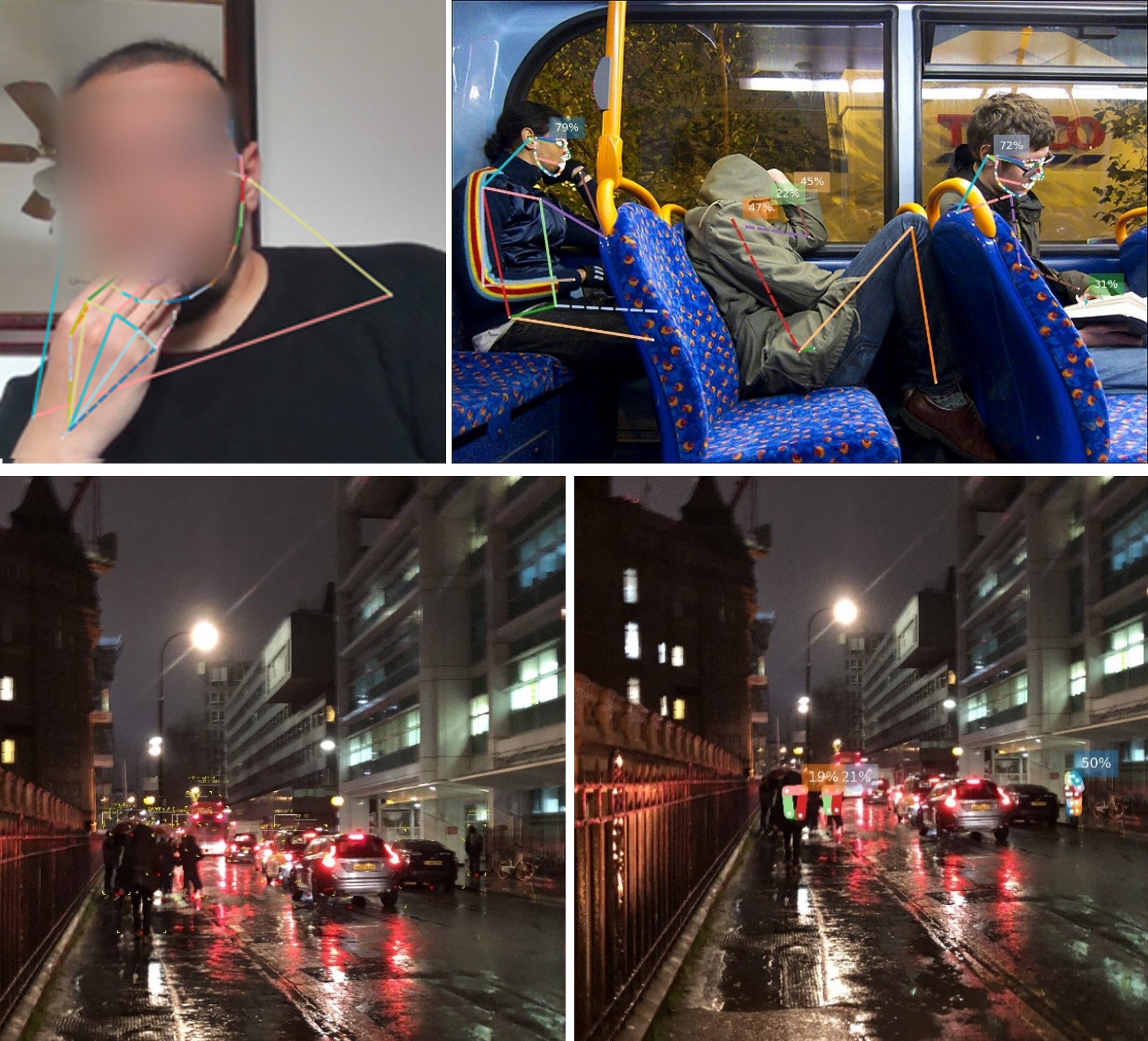}
\caption{{\bf The shortcomings of applying pose estimation as the backbone for FaceTouch. The figure shows several real-world cases under different environmental conditions and complex urban scenes (beyond a single face person).}
}
\label{fig8}
\end{figure}

The proposed method demonstrates originality in the analysis of a broad range of images of face touches that are representative of a variety of human poses. It shows a strong pragmatism when dealing with image and video streams of complex environments with low-resolution of human representations. From images, the FaceTouch framework can detect face touches in real-time, with 25 FPS, in the case of MobileNet. The Facetouch framework shows high performance when it is deployed in real-world settings. This system can also be integrated to assist visually impaired individuals to help them navigate complex urban environments at a safe social distance as presented in \cite{37}, while being aware of others who are touching their faces. Fig \ref{fig7} shows the deployment of the framework in several complex settings, highlighting its effectiveness in differentiating for example between a face touch and the action of drinking water, despite the proximity of one’s hand to the face.  It also shows three scenarios where the utility of the Facetouch framework can be useful: 1) it Shows the performance of the framework in video calls where faces are clear, 2) shows the performance of FaceTouch inside a bus where the resolution is very low and faces are most likely to be partially observed in a very low resolution but the overall human pose may indicate whether there is a face touch, and  3) shows the performance of Facetouch in CCTV cameras whereas the scene includes multiple humans in different poses and scales.

While there are several state-of-the-art methods for extracting human 2D poses based on detecting key points in a human body (i.e. openpifpaf \cite{38}, pifpaf \cite{39}, OpenPose \cite{40}) that can be utilised to measure distances between one’s hands and face as shown in Fig \ref{fig8}, they are less accurate in comparison to object detection, in complex outdoor scenes. Fig \ref{fig8} shows the outcomes of utilising Openpifpaf \cite{38} and an object detector, whereas all humans are detected in case of object detection (5 persons), whereas partial key points of 2 persons only are detected when relying on pose estimation. Accordingly, by utilising pose estimation instead of object detection as a backbone for detecting faces, the introduced framework would have missed detecting humans.  On the other side, the outputs of these methods are inconsistent and spare even when multiple humans are detected in comparison to object detection, which makes it challenging to determine actual facial touch arithmetically due to facial occlusions, and partially seeing humans in a given scene. Accordingly, learning to detect face touches based on the representation of a given RGB image opens the door for utilising the introduced framework in complex scenes produced from low-resolution cameras such as CCTV and bus footage.

\section*{Conclusion}

Understanding unconscious face touches especially in public areas, including indoor spaces or urban environments could help in tracing infectious diseases. This paper aimed to contribute to this issue by providing an autonomous framework that can be deployed in CCTV cameras, known as  FaceTouch, to detect hand-to-face touches in untrimmed video streams. From Images, FaceTouch is trained to detect both human bodies and faces to maximise the ability of the introduced framework to detect face touches despite face occlusion or undetected small faces in complex urban scenes. After detecting whether there is a face touch, the framework ensures individuals’ anonymity by applying facial blur, in case of face detection. The framework is trained on a newly introduced dataset that comprises images extracted from the internet, pedestrian cameras, bus CCTV feeds, and Zoom meetings, ensuring a wide range of utility of the presented framework. After training and validating several encoders’ architectures, the overall framework shows a high validation on the test set.
As for future research, one possible future direction to the presented framework would be utilising the temporal information and sequence of events in video streams when humans touch their faces, or any other objects publicly used in a given scene. To achieve this, the introduced framework of FaceTouch could help in pseudo-labelling sequential frames, alongside extending object detection to include other objects besides humans. 

\section*{Acknowledgment}
This work was supported in part by the PROTECT COVID-19 National Core Study on transmission and environment, managed by the Health and Safety Executive on behalf of HM Government. This work was supported in part by the EPSRC [grant number EP/S026347/1], in part by The Alan Turing Institute under the EPSRC grant EP/N510129/1, in part by The Alan Turing Institute’s Data Centric Engineering Programme under the Lloyd’s Register Foundation grant G0095, in part by The Alan Turing Institute’s Defence and Security Programme, funded by the UK Government, in part by The Alan Turing Institute’s Office of National Statistics Programme, funded by the UK Government and in part by the Hong Kong Innovation and Technology Commission (InnoHK Project CIMDA).

\nolinenumbers

%
%
%


\begin{thebibliography}{10}





\bibitem{1}	
Barroso F, Freedman N, Grand S. Self-Touching, Performance, and Attentional Processes. Percept Mot Skills. 1980;50: 1083–1089. doi:10.2466/pms.1980.50.3c.1083

\bibitem{2}	
Kakaraparthi V, Shao Q, Carver CJ, Pham T, Bui N, Nguyen P, et al. FaceSense: Sensing Face Touch with an Ear-worn System. Proc ACM Interact Mob Wearable Ubiquitous Technol. 2021;5: 1–27. doi:10.1145/3478129

\bibitem{3}	
Chen X “Anthony.” FaceOff: Detecting Face Touching with a Wrist-Worn Accelerometer. arXiv; 2020. Available: http://arxiv.org/abs/2008.01769

\bibitem{4}	
Michelin AM, Korres G, Ba’ara S, Assadi H, Alsuradi H, Sayegh RR, et al. FaceGuard: A Wearable System To Avoid Face Touching. Front Robot AI. 2021;8: 612392. doi:10.3389/frobt.2021.612392

\bibitem{5}	
Ibrahim MR, Haworth J, Cheng T. Understanding cities with machine eyes: A review of deep computer vision in urban analytics. Cities. 2020;96: 102481. doi:10.1016/j.cities.2019.102481

\bibitem{6}	
Ibrahim MR, Haworth J, Christie N, Cheng T, Hailes S. Cycling near misses: a review of the current methods, challenges and the potential of an AI-embedded system. Transp Rev. 2020; 1–25. doi:10.1080/01441647.2020.1840456

\bibitem{7}	
Ibrahim MR, Haworth J, Christie N, Cheng T. CyclingNet: Detecting cycling near misses from video streams in complex urban scenes with deep learning. IET Intell Transp Syst. 2021; itr2.12101. doi:10.1049/itr2.12101

\bibitem{8}	
Ibrahim MR, Haworth J, Cheng T. WeatherNet: Recognising Weather and Visual Conditions from Street-Level Images Using Deep Residual Learning. ISPRS Int J Geo-Inf. 2019;8: 549. doi:10.3390/ijgi8120549

\bibitem{9}	
Ibrahim MR, Haworth J, Cheng T. URBAN-i: From urban scenes to mapping slums, transport modes, and pedestrians in cities using deep learning and computer vision. Environ Plan B Urban Anal City Sci. 2019; 239980831984651. doi:10.1177/2399808319846517

\bibitem{10}	
Nojavanasghari B, Hughes CE, Baltrusaitis T, Morency L. Hand2Face: Automatic Synthesis and Recognition of Hand Over Face Occlusions. arXiv; 2017. Available: http://arxiv.org/abs/1708.00370

\bibitem{11}	
Bai C, Chen Y-P, Wolach A, Anthony L, Mardini MT. Using Smartwatches to Detect Face Touching. Sensors. 2021;21: 6528. doi:10.3390/s21196528

\bibitem{12}	
Bazarevsky V, Kartynnik Y, Vakunov A, Raveendran K, Grundmann M. BlazeFace: Sub-millisecond Neural Face Detection on Mobile GPUs. arXiv; 2019. Available: http://arxiv.org/abs/1907.05047

\bibitem{13}	
He Y, Xu D, Wu L, Jian M, Xiang S, Pan C. LFFD: A Light and Fast Face Detector for Edge Devices. arXiv; 2019. Available: http://arxiv.org/abs/1904.10633

\bibitem{14}	
Zhang K, Zhang Z, Li Z, Qiao Y. Joint Face Detection and Alignment Using Multitask Cascaded Convolutional Networks. IEEE Signal Process Lett. 2016;23: 1499–1503. doi:10.1109/LSP.2016.2603342

\bibitem{15}	
Deng J, Guo J, Zhou Y, Yu J, Kotsia I, Zafeiriou S. RetinaFace: Single-stage Dense Face Localisation in the Wild. arXiv; 2019. Available: http://arxiv.org/abs/1905.00641

\bibitem{16}	
Hu P, Ramanan D. Finding Tiny Faces. arXiv; 2017. Available: http://arxiv.org/abs/1612.04402

\bibitem{17}	
Yang Y, Song X. Research on Face Intelligent Perception Technology Integrating Deep Learning under Different Illumination Intensities. J Comput Cogn Eng. 2022;1: 32–36. 

\bibitem{18}	
Adiguna R, Soelistio YE. CNN Based Posture-Free Hand Detection. 2018 10th International Conference on Information Technology and Electrical Engineering (ICITEE). Kuta: IEEE; 2018. pp. 276–279. doi:10.1109/ICITEED.2018.8534743

\bibitem{19}	
Liu D, Zhang L, Luo T, Tao L, Wu Y. Towards Interpretable and Robust Hand Detection via Pixel-wise Prediction. 2020. doi:10.1016/j.patcog.2020.107202

\bibitem{20}	
Yang L, Qi Z, Liu Z, Zhou S, Zhang Y, Liu H, et al. A Light CNN based Method for Hand Detection and Orientation Estimation. 2018 24th International Conference on Pattern Recognition (ICPR). Beijing: IEEE; 2018. pp. 2050–2055. doi:10.1109/ICPR.2018.8545493

\bibitem{21}	
Xu C, Cai W, Li Y, Zhou J, Wei L. Accurate Hand Detection from Single-Color Images by Reconstructing Hand Appearances. Sensors. 2019;20: 192. doi:10.3390/s20010192

\bibitem{22}
Kourbane I, Genc Y. Skeleton-aware multi-scale heatmap regression for 2D hand pose estimation. arXiv; 2021. Available: http://arxiv.org/abs/2105.10904

\bibitem{23}
Mishra P, Sarawadekar K. Anchors Based Method for Fingertips Position Estimation from a Monocular RGB Image using Deep Neural Network. arXiv; 2020. Available: http://arxiv.org/abs/2005.01351

\bibitem{24}
Sen A, Mishra TK, Dash R. Design of Human Machine Interface through vision-based low-cost Hand Gesture Recognition system based on deep CNN. arXiv; 2022. Available: http://arxiv.org/abs/2207.03112

\bibitem{25}
Xie H, Wang J, Shao B, Gu J, Li M. LE-HGR: A Lightweight and Efficient RGB-based Online Gesture Recognition Network for Embedded AR Devices. 2019 IEEE International Symposium on Mixed and Augmented Reality Adjunct (ISMAR-Adjunct). 2019. pp. 274–279. doi:10.1109/ISMAR-Adjunct.2019.00-30

\bibitem{26}
Khosla P, Teterwak P, Wang C, Sarna A, Tian Y, Isola P, et al. Supervised Contrastive Learning. Vancouver, Canada.; 2020. Available: http://arxiv.org/abs/2004.11362

\bibitem{27}
Hoffer E, Ailon N. Deep metric learning using Triplet network. 2015. Available: http://arxiv.org/abs/1412.6622

\bibitem{28}
Weinberger KQ, Saul LK. Distance Metric Learning for Large Margin Nearest Neighbor Classiﬁcation. J Mach Learn Res. 2009;10: 207–244. 

\bibitem{29}
Everingham M, Eslami SMA, Van Gool L, Williams CKI, Winn J, Zisserman A. The Pascal Visual Object Classes Challenge: A Retrospective. Int J Comput Vis. 2015;111: 98–136. doi:10.1007/s11263-014-0733-5

\bibitem{30}
Viola P, Jones M. Rapid object detection using a boosted cascade of simple features. Proceedings of the 2001 IEEE Computer Society Conference on Computer Vision and Pattern Recognition CVPR 2001. Kauai, HI, USA: IEEE Comput. Soc; 2001. p. I-511-I–518. doi:10.1109/CVPR.2001.990517

\bibitem{31}
He K, Zhang X, Ren S, Sun J. Deep Residual Learning for Image Recognition. 2016 IEEE Conference on Computer Vision and Pattern Recognition (CVPR). Las Vegas, NV, USA: IEEE; 2016. pp. 770–778. doi:10.1109/CVPR.2016.90

\bibitem{32}
Howard AG, Zhu M, Chen B, Kalenichenko D, Wang W, Weyand T, et al. MobileNets: Efficient Convolutional Neural Networks for Mobile Vision Applications. arXiv; 2017. Available: http://arxiv.org/abs/1704.04861

\bibitem{33}
Sandler M, Howard A, Zhu M, Zhmoginov A, Chen L-C. MobileNetV2: Inverted Residuals and Linear Bottlenecks. 2018 IEEE/CVF Conference on Computer Vision and Pattern Recognition. Salt Lake City, UT: IEEE; 2018. pp. 4510–4520. doi:10.1109/CVPR.2018.00474

\bibitem{34}
Ibrahim MR, Lyons T. ImageSig: A signature transform for ultra-lightweight image recognition. 2022 IEEE/CVF Conference on Computer Vision and Pattern Recognition Workshops (CVPRW). New Orleans, LA, USA: IEEE; 2022. pp. 3648–3658. doi:10.1109/CVPRW56347.2022.00409

\bibitem{35}
Dosovitskiy A, Beyer L, Kolesnikov A, Weissenborn D, Zhai X, Unterthiner T, et al. An Image is Worth 16x16 Words: Transformers for Image Recognition at Scale. arXiv; 2021. Available: http://arxiv.org/abs/2010.11929

\bibitem{36}
Selvaraju RR, Cogswell M, Das A, Vedantam R, Parikh D, Batra D. Grad-CAM: Visual Explanations From Deep Networks via Gradient-Based Localization. International Conference on Computer Vision (ICCV). 2017. p. 9. 

\bibitem{37}
Martinez M, Yang K, Constantinescu A, Stiefelhagen R. Helping the Blind to Get through COVID-19: Social Distancing Assistant Using Real-Time Semantic Segmentation on RGB-D Video. Sensors. 2020;20: 5202. doi:10.3390/s20185202

\bibitem{38} 
Kreiss S, Bertoni L, Alahi A. OpenPifPaf: Composite Fields for Semantic Keypoint Detection and Spatio-Temporal Association. arXiv; 2021. Available: http://arxiv.org/abs/2103.02440

\bibitem{39}
Kreiss S, Bertoni L, Alahi A. PifPaf: Composite Fields for Human Pose Estimation. arXiv; 2019. Available: http://arxiv.org/abs/1903.06593

\bibitem{40}
Cao Z, Hidalgo G, Simon T, Wei S-E, Sheikh Y. OpenPose: Realtime Multi-Person 2D Pose Estimation using Part Affinity Fields. arXiv; 2019. Available: http://arxiv.org/abs/1812.08008



\end{thebibliography}
\end{document}